%% file: main.tex
\documentclass[10pt,letterpaper]{article}
\usepackage{helvet}

\usepackage{graphicx}
\usepackage{caption}
\usepackage{tabularx}
\usepackage{graphics}
\usepackage{wrapfig}
\usepackage[font=scriptsize]{caption}
\usepackage{url}
\usepackage{amsgen}
\usepackage{amsfonts}
\usepackage{amsbsy}
\usepackage{amssymb}
\usepackage{amsmath}
\usepackage{cite}
\usepackage{subfigure,subfigmat}
\usepackage{wrapfig}
\usepackage{epsfig}
\usepackage{margins}
\usepackage{fullpage}
\usepackage{bbding}
\usepackage{algorithm, algorithmic}
\usepackage{url}
\usepackage{multirow}
\usepackage{makecell}
\usepackage{wrapfig}
\usepackage{comment}
\usepackage{color}

\usepackage[symbol]{footmisc}

\usepackage{authblk}

\title{\textbf{\LARGE Intelligent Policing Strategy for Traffic Violation Prevention}}
\author[$^\dagger$]{\large Monireh Dabaghchian}
\author[$^*$]{Amir Alipour-Fanid}
\author[$^*$]{Kai Zeng}
\affil[$^\dagger$]{Department of Computer Science, and Cybersecurity Assurance and Policy (CAP) Center, Morgan State University}
\affil[$^*$]{Department of Electrical and Computer Engineering, George Mason University}
\affil[ ]{\textit {E-mail: $^\dagger$monireh.dabaghchian@morgan.edu, \{$^*$aalipour, $^*$kzeng2\}@gmu.edu}}
 \date{\vspace{-5ex}}

\begin{document}
  \maketitle


\input{abstractone}
\input{Introduction}

\input{Traffic_Violation_Prevention_Model}

\input{Intelligent_Policing_Strategy}

\input{Case_Study}

\input{Conclusion}



\bibliographystyle{IEEEtran}
\bibliography{IEEEabrv,bibfile}
\clearpage

\end{document}

%% file: abstractone.tex
\section{Abstract}
Police officer presence at an intersection discourages 
a potential traffic violator 
from violating the law. 
It also alerts the motorists' consciousness to take precaution and follow the rules. 
However, due to the abundant intersections and shortage of human resources, it is not possible to assign a police officer to every intersection. 
In this paper, we propose an intelligent and optimal policing strategy for traffic violation prevention. 
Our model consists of a specific number of targeted intersections and two police officers with no prior knowledge on the number of the traffic violations in the designated intersections. 
At each time interval, the proposed strategy, assigns the two police officers to different intersections such that at the end of the time horizon, maximum traffic violation prevention is achieved. 
Our proposed methodology adapts the PROLA (Play and Random Observe Learning Algorithm) algorithm \cite{Dab2018} to achieve an optimal traffic violation prevention strategy. 
Finally, we conduct a case study to evaluate and demonstrate the performance of the proposed method.

%% file: Introduction.tex
\section{Introduction and Motivation}


\hspace{4mm} A popular and well-sensible proverb, ``Prevention is better than cure", reveals that efforts to prevent a negative incident from happening are much better than finding out a solution to resolve a problem.
Along with this belief, public law enforcement agencies (Local police, Sheriffs' departments, Special police, Federal and State police) also endeavor in reduction of criminal activities, robbery, assault, major/minor traffic violation, and physical violence through mission assignments or a regular and predefined strategies \cite{JP2017}.
Traffic violation is one of the major concerns for the law enforcement agencies among other illegal and unlawful activities.
In this regard, intersections in city streets or main roads are the critical and important regions in severity and frequency of traffic violation in the records \cite{HW2006,CY2006}.
There are, for example, various intersection traffic violation types, including but not limited to disobeying traffic lights, signs (yield on green, do not turn right, left and U-turn) or signals, blocking or retarding traffic,
failure to stop or yield to pedestrian,
speeding,
improper blowing of horn,
improper turn,
racing, dragging, or contest for speed.
Depending on the violation degree, traffic law violation may cause single/multi-vehicle collision or in some cases fatal pedestrian/vehicle crash.
Law enforcement agencies utilize various effective law enforcement methods such as proper traffic signs deployment, radar speed gun, red light camera and etc, in order to prevent the traffic violations \cite{MG2005}.

Among all approaches, police officer presence in the field is the most effective and functional way to help reduce the traffic violation dramatically. 
Vehicle drivers take precautions and get alerted when observing the police officer or his/her car standing close to the intersections. 
Therefore, traffic managers and engineers assign police officers to some places based on some strategies to monitor and combat traffic violation. 
Typically, these strategies are based on personal experiences i.e., prior knowledge on the frequency of traffic violations happening in a specific region.
Although the value and usefulness of the law enforcement officer's experiences cannot be ignored \cite{FJ2018}, however, research community also continuously strives to discover superior and optimal strategies based on the advanced technologies and scientific methods such as historical data (evidence-based records) analysis methodologies \cite{CR2018, LEADS2018, P2018}.

Nowadays, due to the rapid outward growth and development of the cities and sub-urban areas on the one hand, and shortage of law enforcement resources on the other hand, the need for intelligent and cost-effective policing strategies are of a great significance and importance to the law enforcement policy planners and administration.
Inspired by the idea of police officer presence in the fields for prevention purposes, in this paper, we propose a novel intelligent policing strategy which mathematically is proven to be an optimal strategy subject to the proposed model.

In this model, we consider a geographical region of a city/sub-urban which consists of a set of intersections.
We then consider a police officer which intends to choose and attend a specific intersection out of all intersections at different time intervals to help prevent the traffic violations.
Considering various traffic violation rates for different intersections, the goal is for the police officer to find the intersection with the most traffic violation rates and to be present in that intersection more often than others to maximize the number of traffic violation prevention.
We utilize an online learning strategy called PROLA (Play and Random Observe Learning Algorithm) \cite{Dab2018} for the police officer to learn the traffic violation rates at different intersections.
The proposed method promised to be efficient and cost-effective by enabling intelligent police officer assignment to work actually smarter, not harder.

 
\begin{figure}[t]
		\centering
			\includegraphics[scale=0.67]{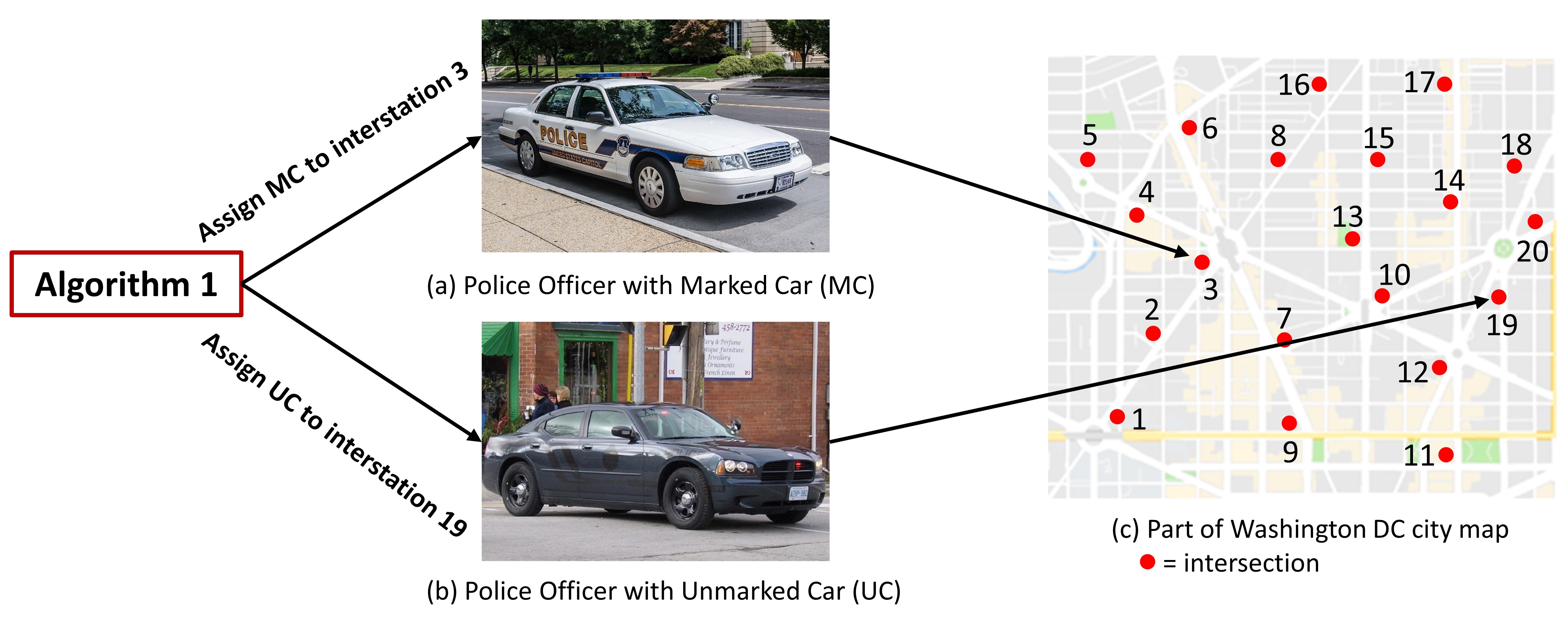}
                    \vspace{-1mm}
			\caption{Traffic violation prevention model}
                \vspace{-4mm}

					\label{modelfig}
\end{figure}

\textbf{Challenge:} 
The key in the police officer's efficient learning process is to be able to observe/know whether his/her presence in a specific location prevented traffic violations or not. 
However, due to the nature of this problem, any time attending a location the police officer will have no idea whether his presence was effective or not.
This is because either there were no violators in the area or if there were any, they recognized the police officer via his marked car and did not violate the traffic law. 
This poses a challenge in obtaining an optimal policing strategy to prevent the traffic violations as based on the existing work in the literature for any agent interacting with the environment, the agent needs to observe some feedback on the action he/she takes. 
However, in this problem, the police officer as an agent is not able to observe the feedback on his action which is defined as the intersection he/she attends.
In order to overcome the challenge, we apply the PROLA algorithm for optimal policing strategy that achieves an optimal performance in terms of traffic violation prevention.

%% file: Traffic_Violation_Prevention_Model.tex
\section{Traffic Violation Prevention Model}

\hspace{4mm} Our model consists of two types of law enforcement officers.
One of the officers wears a police uniform and uses a marked car, so he/she and his/her car are recognizable by the other vehicle drivers.  
We name this type of officer as \emph{Police Officer with Marked Car (POMC)} and for simplicity of the notation we denote it by MC.  
The other officer is not in police uniform and utilizes unmarked police car, so he/she and his/her car are not recognizable as a law enforcement officer by the other vehicle drivers. 
We also name this type of officer as \emph{Police Officer with Unmarked Car (POUC)} and denote it by UC.
We also consider that in the targeted area for traffic violation prevention there are $K$ intersections. 
We assume that MC and UC get assigned to a specific intersection for a fixed time interval. At the end of the time interval, Algorithm 1 is run and the MC and UC are assigned to next chosen intersections.
We also assume that the MC and UC are never assigned to the same intersection. 
Figure \ref{modelfig} illustrates the main components of the proposed model.

%% file: Intelligent_Policing_Strategy.tex
\section{Intelligent Policing Strategy}


\hspace{4mm} Once MC is assigned to an intersection, his/her presence prevents the traffic violation but cannot see the reward.
However, once UC is assigned to an intersection, the drivers are not aware of the presence of the police, so a potential violator may violate the traffic law so then the UC can observe the violator but receive no reward. 
The point is that MC cannot observe the traffic violation prevention (reward), hence, the MC is not aware of his/her presence impact. 
However, UC can observe the traffic violation, but his/her presence cannot prevent the traffic violation (meaning no reward). Instead, he gathers data to learn the intersection(s) that have high traffic violation rates.

In this regard, our proposed method adresses the following question.
\emph{Which intersection the MC and UC have to be assigned in each time interval over the whole time horizon such that the traffic violation due to the presence of the MC is minimized?}
Since the MC needs to learn and prevent traffic violation at the same time and it has no prior knowledge of the violation activity on different intersections beforehand, we formulate this problem as an online learning problem as follows.

\begin{algorithm}[t]
\caption*{\textbf{Algorithm 1: PROLA (Play and Random Observe Learning Algorithm) \cite{Dab2018} }}
    \label{alg:optimizatiomn}
 \begin{algorithmic}
 {
 \STATE \textbf{Parameters:}  $\gamma \in \left( 0, 1\right)$,   
$\eta \in  \left( 0,   \frac{\gamma}{2(K-1)}                 \right]$. \\

\textbf{Initialization:} $\omega_1(i) = 1$, \quad $i = 1,...,K.$

\textbf{For each} $t = 1,2,...,T$ \\

\quad\quad 1. Set
$p_t(i)= (1-\gamma ) \frac{\omega_t(i)}{\sum_{j=1}^K\omega_t(j)} + \frac{\gamma}{K}$, \quad\quad $i=1,...,K$.  \\

\hangindent=2.4em
\hangafter=1
\quad\quad 2. MC select intersection $I_t \sim\ p_t$ and accumulate the unobservable reward $x_t(I_t) $. \\

\hangindent=2.4em
\hangafter=1
\quad\quad 3. UC chooses an intersection $J_t$ other than the intersection chosen by the MC uniformly at random\\
\hspace{10mm} and observes its reward $x_t(J_t)$ based on equation (\ref{rwrdDef}).\\

\quad\quad 4. For $j = 1,...,K$ 
\[
{\hat{x}_t}(j)  =
\begin{cases}
\frac{x_t(j)}{(1/(K-1))(1 - p_t(j))}, & j = J_t \\
0, & o.w.,
\end{cases}
\]
\hspace{48mm}
$\omega_{t+1}(j)=\omega_t(j) \exp (\eta \hat{x}_t(j))$.\\
   }
\end{algorithmic}
\end{algorithm}


\begin{table}
\centering
\caption{Main Notation}
\begin{tabular}{|c|l|} \hline
$T$& total number of time intervals  \\ 
$K$        & total number of intersections    \\ 
$I_t$      & index of the intersection that MC is assigned at time $t$   \\ 
$J_t$      & index of the intersection that UC is assigned at time $t$  \\ 
$R $       & total regret of the MC  \\ 
$\gamma$   &exploration rate        \\ 
$\eta$     &learning rate    \\ 
$p_t\left(i\right)$   & MC assignment distribution on intersection $i$ at time $t$      \\ 
$\omega_t\left(i\right)$   & weight assignment to intersection $i$ at time $t$ \\ \hline
\end{tabular}
\end{table}

 
\begin{figure*}[t]
		\centering
			\includegraphics[scale=0.76]{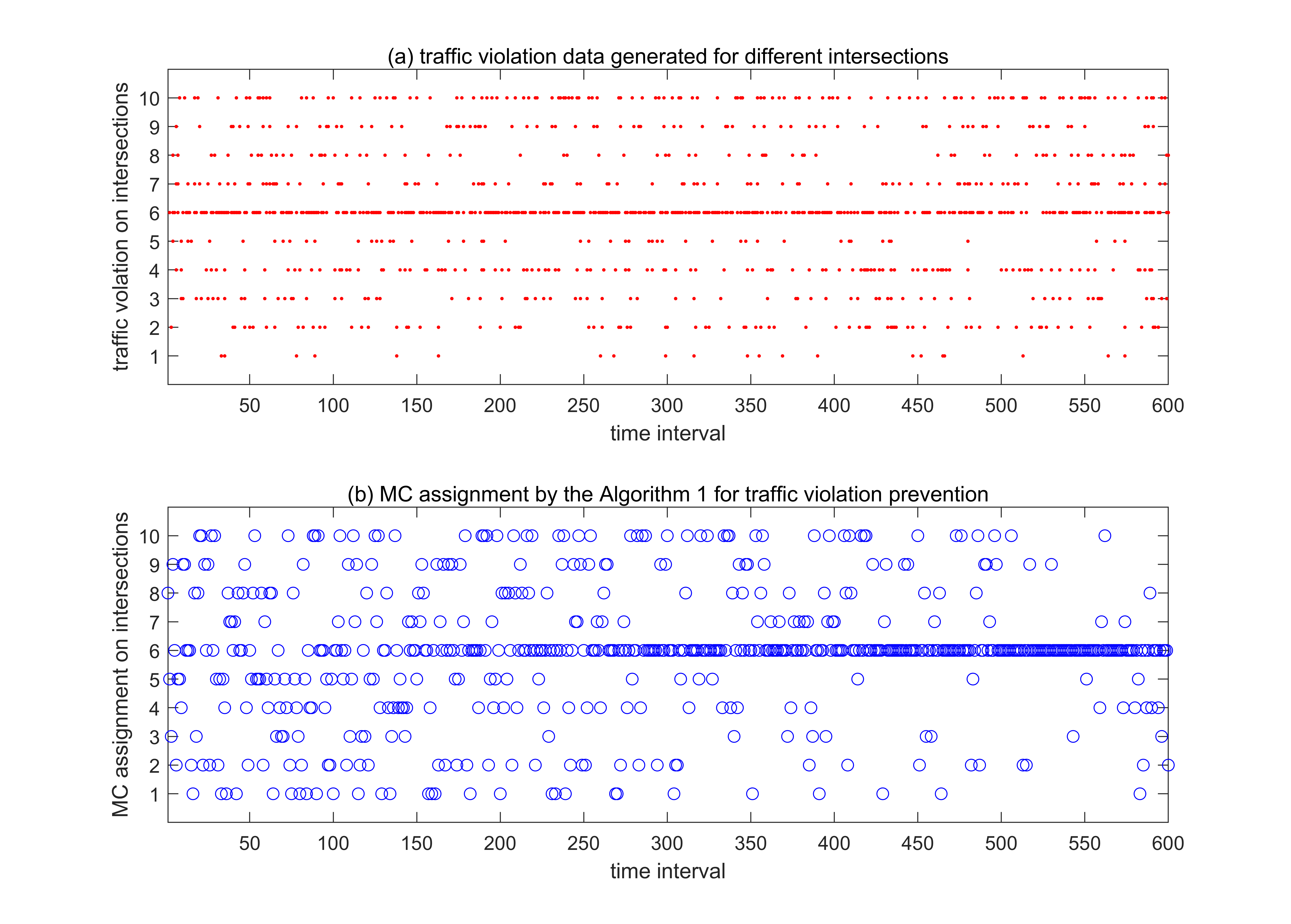}
			\caption{Algorithm 1 performance on traffic violation prevention by MC assignment}
					\label{VP}
\end{figure*}
We define $x_t(j)$ as the MC's reward on the intersection $j$ ($ 1 \leq j \leq K $) at time interval $t$ ($1\leq t\leq T$) where $K$ is the number of the intersections in the region and $T$ is the total number of time intervals.
Table 1 summarizes the main notation used in this paper.

Without loss of generality, we normalize $x_t(j)\in\left[0,1\right]$. 
More specifically:
\begin{align}
\label{rwrdDef}
x_t(j) =
\begin{cases}
1, & \text{\emph{traffic is violated at intersection $j$ at time interval $t$}} \\
0, & o.w.
\end{cases}.
\end{align}

Suppose the proposed algorithm applies learning policy $\varphi$ to assign MC and UC to the intersections.
The aggregated expected reward of the MC by time interval $T$ is equal to
\begin{equation}
	\label{equ6}
				{G_\varphi(T)} = \bold{E}_\varphi\left[\sum\limits_{t=1}^T x_t(I_t)\right], \\
\end{equation}		
where $I_t$ indicates the intersection index that the MC is assigned by the algorithm at time $t$. 
The MC's goal is to maximize the expected value of the aggregated reward, thus to minimize the traffic vilolation,
\begin{equation}
\label{equ7}
\mathrm{maximize} \quad G_{\varphi}(T).
\end{equation}

For a learning algorithm, regret is commonly used to measure its performance \cite{Dab2018}.
The regret of the MC can be defined as follows:
\begin{equation}
\label{equ8}
Regret =  G_{max} - G_{\varphi}\left(T\right),
\end{equation}
where
\begin{equation}
G_{max} = \max\limits_j \sum\limits_{t=1}^T x_{t}\left(j\right).
\end{equation}

The regret measures the gap between the accumulated reward achieved by applying a learning algorithm and the maximum accumulated reward the MC can obtain when it always (i.e., for the all time intervals) stays in the intersection with the highest number of traffic violations. 
From the MC point of view, we call this specific intersection as \emph{Best Intersection (BI)}.
The BI is the intersection with highest accumulated reward up to time $T$. 
Then, the problem can be transformed to minimize the regret as follows:
\begin{equation}
\label{equ10}
\mathrm{minimize} \quad G_{max} - G_{\varphi}\left(T\right).
\end{equation}

In order to solve the Eq. (6), we adapt the \emph{PROLA (Play and Random Observe Learning Algorithm)} algorithm \cite{Dab2018}, and propose MC and UC to be assigned to different intersections over the time based on the strategy outlined in Algorithm 1.
We have proved the optimality of this algorithm in \cite{Dab2018}.

%% file: Case_Study.tex
\section{Case Study and Performance Evaluation}

\hspace{4mm} Number of intersections in the targeted area may vary and depend on several urban design factors such as whether it is a residential, commercial or industrial area, etc.
In this case study, we consider various scenarios with different number of intersections i.e., $K \in \{10,20,30,40,50\}$.  
We can also assume that each time interval is one hour and at the beginning of each time interval, police officers (MC/UC) are assigned to different intersections according to Algorithm 1.

At the end of each time interval, UC reports the number of the traffic violations (if ever happened) in the intersection to the algorithm.
Then, based on the information acquired till the end of that time interval, the algorithm chooses the indices of the two intersections for the next time interval in which both officers will be assigned to them.
This process is repeated for the whole time horizon $T$.



 \begin{figure*}
  \subfigure[MC assignment probability on the intersections,  $K=10$.]{\includegraphics[scale=0.52]{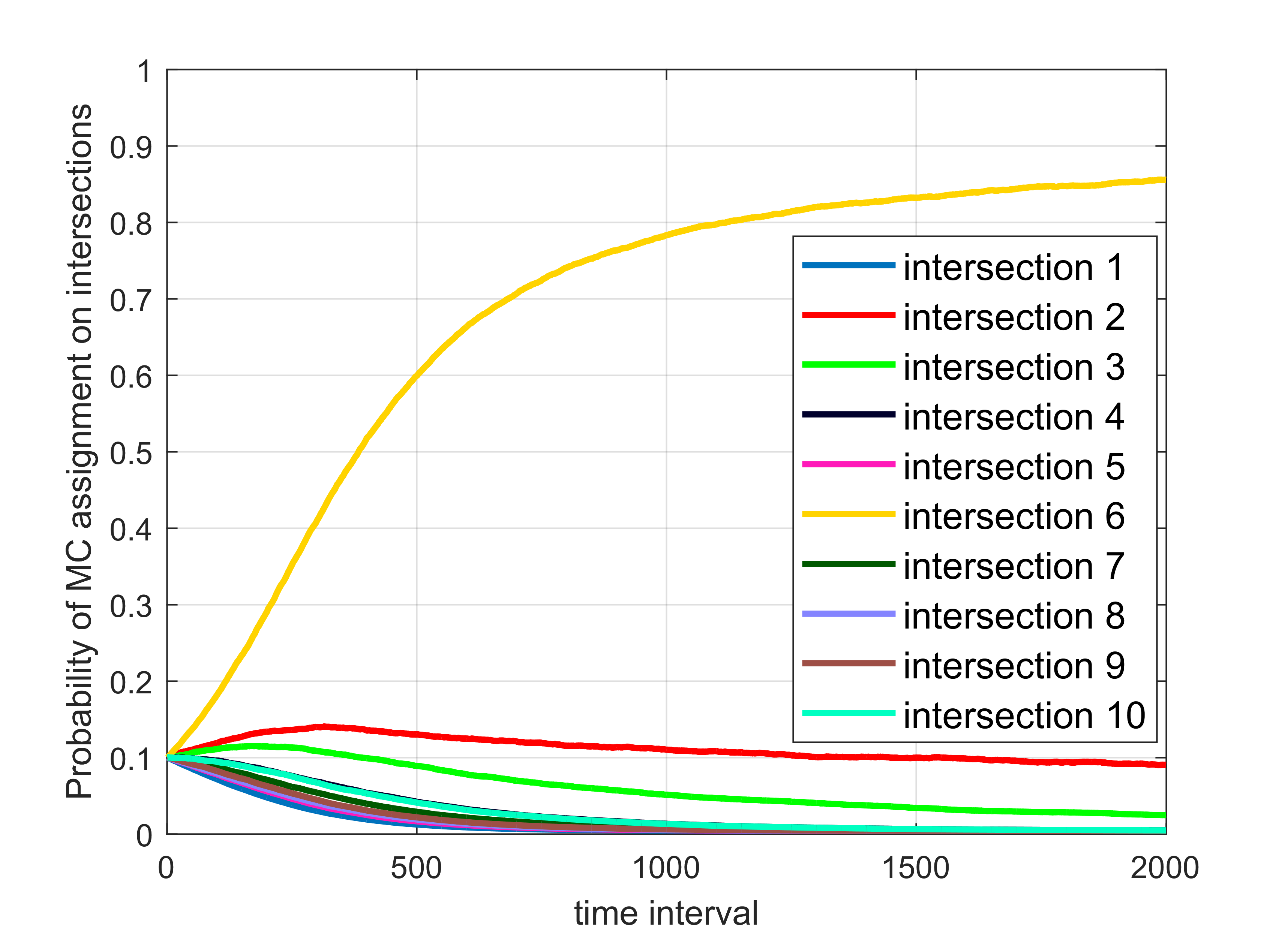}}\hspace{1.5cm}
  \subfigure[Total regret of MC for different K over the time.]{\includegraphics[scale=0.52]{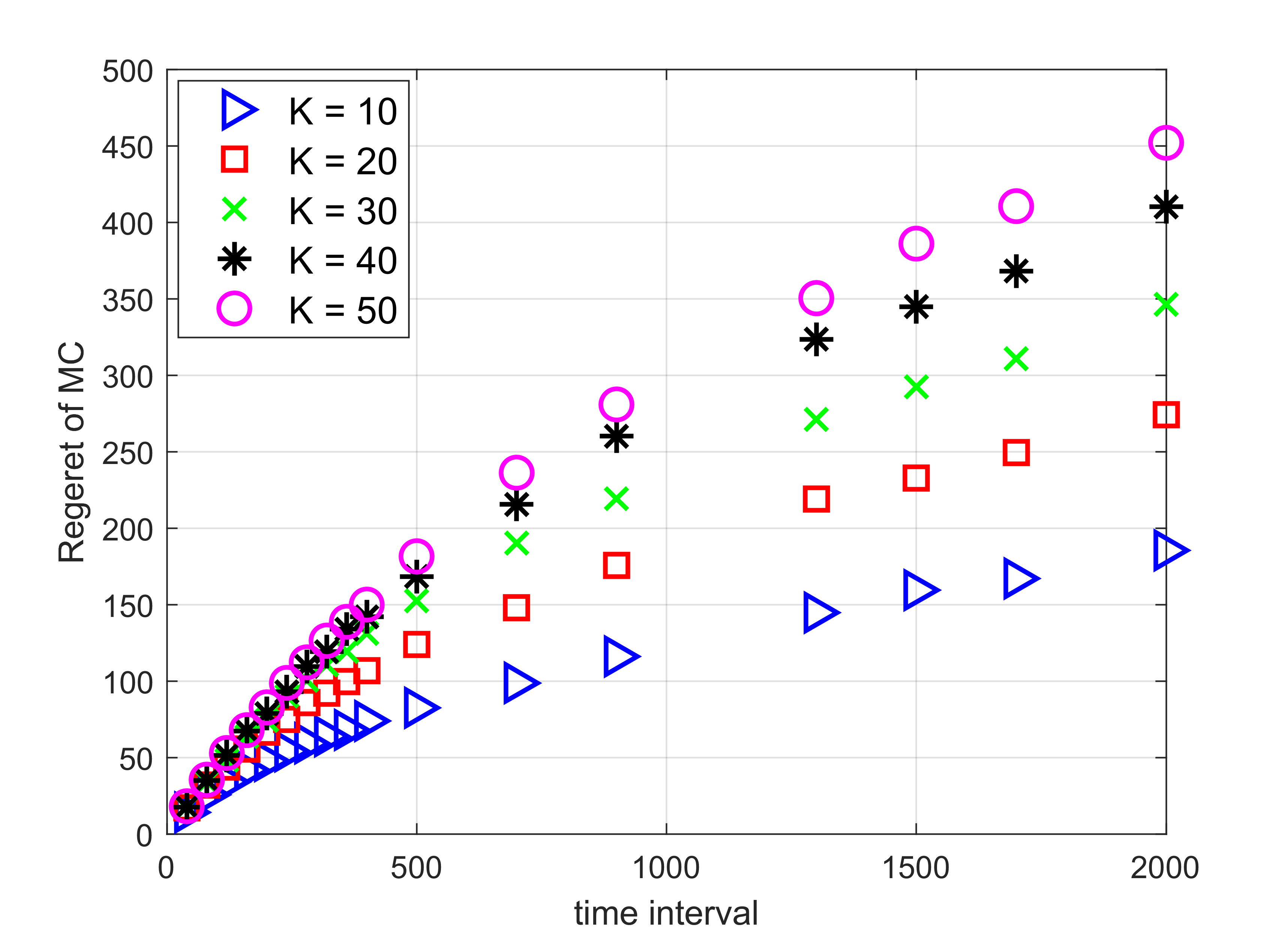}}
	  \caption{Algorithm 1 performance evaluation}
				\label{traintestfig}
\end{figure*}

We assume that at each time interval, traffic violation follows a Bernoulli distribution.
For instance, for $K=10$ we assume $p_1 = 0.04, p_2 = 0.2, p_3=0.17,p_4=0.2,p_5=0.08, p_6 =0.6, p_7=0.16,  p_8=0.1, p_9= 0.12, p_{10}=0.2$, where $p_i$ for $i=1,2,...,10$ indicates the probability of the traffic violation at each time interval in the  $i$th intersection.
Figure \ref{VP}(a) illustrates the generated traffic violation for different intersections in time horizon $1$ to $600$ according to the aforementioned distribution. 
According to the BI definition, in the generated data, intersection 6 is the BI.
Figure \ref{VP}(b) shows how Algorithm 1 assigns MC to different intersections to maximize the traffic violation prevention. 
It is shown in this figure that as time goes on, the MC is assigned to intersection 6 (i.e., BI) more frequently to prevent traffic violation.  

For the case of $K=10$, Fig. 3(a) illustrates the probability of assigning the MC to the various intersections by the Algorithm 1 for the whole time horizon $T$. 
Intersection 6 is the BI, so the probability that the Algorithm 1 assigns MC to the BI increases as time goes on. 
Figure 3(b) shows the MC's regret for various number of intersections.
As the number of the intersections increases the MC's regret increases, as well. 
However, since the nature of PROLA algorithm is a  \emph{no-regret} algorithm \cite{Dab2018, AB2007}, the MC's regret independent from $K$ approaches to zero as $T \xrightarrow{} \infty$.

\emph{\textbf{Remark:} Since the proposed learning algorithm performs online, if the BI changes to another intersection, the algorithm adaptively learns and converges to the new BI. As a result, MC is assigned to the new BI more frequently to maximize the traffic violation prevention.}

%% file: Conclusion.tex
\section{Conclusion}
\hspace{4mm} In this paper, we proposed an intelligent policing strategy for traffic violation prevention. 
The proposed online learning algorithm assigns two types of police officers, Police Officer with Marked Car (MC) and Unmarked Car (UC),  to various intersections strategically such that at the end of the time horizon maximum traffic violation prevention is achieved.
We evaluated the performance of the algorithm through simulation on various settings and showed that as time goes on, the algorithm learns the environment and assigns the MC to the best intersection in the region to maximize the traffic violation prevention.

We hope that the proposed methodology and the model insight can further pave the path to invent new types of intelligent algorithms which are essential to unlock the full potential of artificial intelligence in strategic policing.
